\documentclass[10pt,journal,compsoc]{IEEEtran}
\usepackage{balance}

\ifCLASSOPTIONcompsoc
  \usepackage[nocompress]{cite}
\else
  \usepackage{cite}
\fi

\usepackage{amsmath}
\usepackage{amssymb}
\usepackage{xcolor}
\usepackage{colortbl}
\usepackage{float}
\usepackage{algorithm}
\usepackage{algorithmic}
\usepackage{graphicx}
\usepackage{subfig}
\usepackage{booktabs}
\usepackage{url}

\DeclareUnicodeCharacter{2061}{}

\begin{document}

\title{Pressure Ulcer Categorisation using Deep Learning: A Clinical Trial to Evaluate Model Performance}

\author{\IEEEauthorblockN{Paul Fergus\IEEEauthorrefmark{1}, Carl Chalmers\IEEEauthorrefmark{1}, William Henderson\IEEEauthorrefmark{2}, Danny Roberts\IEEEauthorrefmark{2}, and Atif Waraich\IEEEauthorrefmark{1}.
}
\IEEEauthorblockA{\\ \IEEEauthorrefmark{1}Liverpool John Moores University, Byrom Street, Liverpool, L3 3AF, UK}
\IEEEauthorblockA{\\ \IEEEauthorrefmark{2}Mersey Care NHS Foundation Trust, V7 Building, Prescot, L34 1PJ, UK}
\IEEEcompsocitemizethanks{
	\IEEEcompsocthanksitem Paul Fergus, Carl Chalmers and Atif Waraich are with the Faculty of Engineering and Technology at Liverpool John Moores University, Byrom Street, Liverpool, L3 3AF, UK. (E-mail: p.fergus@ljmu.ac.uk, c.chalmers@ljmu.ac.uk, a.waraich@ljmu.ac.uk)\protect\\
}
\IEEEcompsocitemizethanks{
	\IEEEcompsocthanksitem William Henderson and Danny Roberts are at Mersey Care NHS Foundation Trust, V7 Building, Prescot, L34 1PJ, UK. (E-mail: william.henderson@merseycare.nhs.uk, danny.roberts@merseycare.nhs.uk)\protect\\
}}

\markboth{}%
{Fergus \MakeLowercase{\textit{et al.}}: Bare Demo of IEEEtran.cls for Computer Society Journals}

\IEEEtitleabstractindextext{%
\begin{abstract}
Pressure ulcers are a challenge for patients and healthcare professionals. In the UK, 700,000 people are affected by pressure ulcers each year. Treating them costs the National Health Service £3.8 million every day. Their etiology is complex and multifactorial. However, evidence has shown a strong link between old age, disease-related sedentary lifestyles and unhealthy eating habits. Pressure ulcers are caused by direct skin contact with a bed or chair without frequent position changes. Urinary and faecal incontinence, diabetes, and injuries that restrict body position and nutrition are also known risk factors. Guidelines and treatments exist but their implementation and success vary across different healthcare settings. This is primarily because healthcare practitioners have a) minimal experience in dealing with pressure ulcers, and b) a general lack of understanding of pressure ulcer treatments. Poorly managed, pressure ulcers lead to severe pain, poor quality of life, and significant healthcare costs. In this paper, we report the findings of a clinical trial conducted by Mersey Care NHS Foundation Trust that evaluated the performance of a faster region-based convolutional neural network and mobile platform that categorised and documented pressure ulcers. The neural network classifies category I, II, III, and IV pressure ulcers, deep tissue injuries, and unstageable pressure ulcers. Photographs of pressure ulcers taken by district nurses are transmitted over 4/5G communications to an inferencing server for classification. Classified images are stored and reviewed to assess the model’s predictions and relevance as a tool for clinical decision making and standardised reporting. The results from the study are encouraging and show that using 216 images collected over the eight-month trial it was possible to generate a mean average Precision=0.6796, Recall=0.6997, F1-Score=0.6786 with 45 false positives using an @.75 confidence score threshold.     
\end{abstract}
}

\maketitle

\IEEEdisplaynontitleabstractindextext

\IEEEpeerreviewmaketitle

\IEEEraisesectionheading{\section{Introduction}\label{sec:introduction}}

\IEEEPARstart {I}{n} the UK, 700,000 people are affected by pressure ulcers each year \cite{stephens2018understanding}. According to National Health Service (NHS) Improvement, pressure ulcers cost the NHS more than £3.8 million every day to manage and treat \cite{NHSImprovement2018}. In England, 24,674 patients developed a new pressure ulcer between April 2015 and March 2016 \cite{NHSImprovement2018}. UK-wide, the number of new pressure ulcers in 2017/2018 was estimated to be 200,000. The cost of treating a category I pressure ulcer is equal to £1,124 while a category IV costs £14,108. \cite{dealey2012cost}, \cite{guest2018cohort}. A House of Lords strategy discussion group in the UK in November 2017 reported that £5 billion is spent on wound care every year - a similar financial cost to the NHS for managing obesity \cite{browning2017house}. Malpractice claims against UK trusts relating to pressure ulcers increased by 43\% in the three years leading up to 2017-18. The number of litigation cases increased from 279 in 2014-15 to 399 in 2017-18 with the bill to the NHS increasing 53\% from more than £13.6m to £20.8m. In total, more than £72.4m was spent on pressure ulcer claims over that period. While most cases are settled out of court for approximately £20-30,000, some have cost the NHS as much as £3m \cite{white2015pressure}.  
\par
Pressure Ulcers are caused by unrelieved pressure over bony parts of the body \cite{avsar2021dressings}. Skin shearing, friction, moisture and faecal soiling increase the risk of pressure ulcers significantly, these conditions are common in patients that are elderly, sick, debilitated or paralysed \cite{woo2017management}. Poorly managed, pressure ulcers can lead to severe pain, reduced quality of life and significant economic costs to the NHS \cite{guest2015health}. Pressure ulcers are diagnosed as either a category I, II, III, IV pressure ulcer, a deep tissue injury (DTI) or unstageable. Pressure ulcers often occur on a) the ischial region (buttocks) typical for chair-bound patients, b) the back of the heal - in the supine position, c) the sacrum - in the supine position and d) the trochanteric region - in the lateral position \cite{sato2018factors}. A category I diagnosis is given when the surface of the skin is intact but reacts to injury by becoming red and hyperaemic. Category II ulcers occur in the epidermis and dermis layers where they can become necrotic and cause skin cover deficiency. Category III ulcers involve subcutaneous tissue and category IVs have lesions that penetrate underlying muscle or bone. Category III and IV ulcers often have large amounts of necrotic tissue deep within the wound cavity. DTIs appear underneath intact skin and present themselves as deep bruises, which can deteriorate into a deep pressure ulcer. Unstageable wounds have an undetermined level of tissue damage covered with slough or eschar/necrotic tissue. Unstageable pressure ulcers can only be categorised once they have been debrided. 
\par
Guidelines for pressure ulcer risk assessment and prevention are provided by the National Institute for Care Excellence (NICE) \cite{stansby2014prevention}. The NHS Safety Thermometer incident reporting system and the Strategic Executive Information System are also widely used across the NHS to allow pressure ulcer incidents to be documented \cite{smith2016pressure}. However, there is significant variation in their implementation \cite{mccaughan2018patients}, \cite{lumbers2019overview}. This is primarily because healthcare practitioners have a) varied experience in dealing with pressure ulcers, and b) a general lack of understanding of pressure ulcers and the treatment thereof \cite{cho2011exploring}. The challenge is to provide a decision-support tool for healthcare practitioners that standardises pressure ulcer categorisation and makes pressure ulcer management more accessible to a wider group of healthcare professionals.  
\par
To address this challenge, we present a pressure ulcer management system that utilises a Faster Region-based Convolutional Neural Network (Faster R-CNN) \cite{ren2016faster} and a mobile platform capable of automatically categorising and reporting pressure ulcers in a standardised manner. The Faster R-CNN is an Artificial Intelligence (AI) algorithm that uses deep learning (DL) and advanced computer vision to detect category I, II, III, and IV pressure ulcers, DTIs, and unstageable pressure ulcers. The system generates results in near real-time. It is not designed to replace human assessments but enhance clinical practice, prevent diagnostic errors and standardise how pressure ulcers are analysed and reported. To the best of our knowledge, the system is the first of its kind to be evaluated in a clinical trial, undertaken by Mersey Care NHS Foundation Trust in the UK.
\par
The remainder of the paper is structured as follows. A discussion on automated image analysis is presented in Section 2. Section 3 describes the proposed methodology and study protocol before the results are presented in Section 4 and discussed in Section 5. The paper is concluded and future work is presented in Section 6.
\section{Automated Image Analysis}
Automated medical image analysis has been an active area of research since scans were digitised and processed with computers. Between the 1970s and 1990s image analysis was performed using edge and line detector filters. For example, snakes active contour models (ACM) were used to perform segmentation in \cite{kass1988snakes}. Later, this approach was extended and used in leg ulcer studies with piecewise B-spline arcs to adaptively initialise the ACMs \cite{jones2000active}. 
\par
Region-based approaches, also known as similarity-based segmentation, appeared in the late 1990s. Both \cite{bertelli2008variational} and \cite{veredas2015efficient} utilised similarity-based segmentation to build colour histogram models, and using Bayesian inference were able to compute the posterior membership probability of pixels belonging to segments in a pressure ulcer image. By assigning pixels to different segments, the ulcer could be deconstructed to measure the wound and its constituent tissue.  
\par
Other image processing approaches include a) spectral clustering \cite{shi2000normalized} which finds segments in images using morphological operators \cite{dhane2016spectral}, b) relationship modelling between density and pixel intensity using synthetic frequencies extracted with contrast changes and energy density models \cite{ortiz2017pressure}, and c) toroidal geometry, where images over multiple contrast levels and varying synthetic frequencies are segmented with the method described in \cite{otsu1979threshold}. 
\par
The focus moved from 2D to 3D image processing in the late 1990s with the introduction of the Measurement of Area and Volume Instrument System (MAVIS) \cite{ plassmann1998mavis}. MAVIS constructs three-dimensional mappings of pressure ulcers by projecting parallel stripes of alternating colours onto the region of interest. The volume of the ulcer is then computed using cubic spline interpolation. Similar 3D image processing approaches were proposed in \cite{albouy20073d} and \cite{yee2016quantitative} where 3D models of wounds were generated by matching calibrated images captured from different angles and Stereoscopic 3D reconstruction.  These approaches work well in facilities with specialised equipment but are impractical in frontline service delivery due to the need for costly and complex lighting, specialised devices, and staff trained to use the systems.
\par
In the 1990s, machine learning algorithms were developed to perform semantic segmentation, data fitting, and statistical classification using image-specific features \cite{gunsel1998temporal}, \cite{pieczynski1992statistical}. Applications were primarily in the medical domain and the feature extraction aspects of the pipeline were manually designed by humans \cite{pham2000current}. Today, features in images are automatically extracted using Convolutional Neural Networks (CNNs) \cite{wang2012machine}, \cite{maier2019gentle}. In fact, since AlexNet (a CNN architecture) \cite{ krizhevsky2012imagenet}, most traditional image processing approaches have been replaced with DL given their ability to solve complex image processing problems. 
\par
CNNs are now widely used to analyse medical images. In pressure ulcer studies there are some notable works. For example, \cite{zahia2018tissue} proposed a system that classifies tissue types and performs segmentation using CNNs. The major limitation however is that low-quality images are used which have limited utility in complex wound analysis. Furthermore, the approach does not consider categorisation. Alongside the need for high-quality images, CNNs require lots of them for successful convergence. Medetec is the most comprehensive open-source pressure ulcer dataset which contains 175 images of pressure ulcers - an insufficient number for training CNN models. Transfer learning can help overcome this issue by using models trained on a large corpus of images and then fine-tuning them with images contained in smaller datasets. For example, in \cite{wang2015unified} Wang et al. used transfer learning and a small dataset of images to train a CNN model to segment wounds and detect infection which produced reasonably good results. Similar results are reported in \cite{li2018composite}, \cite{zhao2019fine} and \cite{chino2020segmenting}. The major limitation is that none of these approaches scale up or provide a clear assessment of their utility in complex wound analysis or clinical practice. 
\par
There are several limitations with current approaches a) they use low-resolution images given that there are no publicly available high-resolution images of pressure ulcers and b) they do no not focus on the full compliment of pressure ulcer categories or tissue types. Any viable pressure ulcer tool must be able to classify all pressure ulcer categories. This requirement must preclude any use of segmentation or measurement if a full assessment of pressure ulcer wounds is to be achieved, which in most studies, is not the case.
\par  
In this study, we build on the state-of-the-art and propose a DL system that can categories category I, II, III, IV, DTI and Unstageable pressure ulcers. The approach is evaluated in a clinical trial to determine if it can provide the intended benefits to healthcare practitioners as a tool for categorising and reporting pressure ulcers in a standardised way.  
\section{Methodology}
This section describes the data collection strategy used in this study. A discussion is also presented on how a custom dataset, image augmentation, and transfer learning are used to train a Faster R-CNN model to categorise pressure ulcers. The model is integrated into a mobile platform for use in a clinical trial which is also discussed. The section is concluded with a set of evaluation metrics for model training and inference conducted in the clinical trial.
\subsection{Data Collection and Pre-processing}
The Medetec pressure ulcer dataset provides a baseline image set in this study which contains 174 images of pressure ulcers (classes included are category I, II, III, and IV, DTI and unstageable). The dataset is supplemented with and additional 675 images (across the same classes) acquired from Google. Images were used based on the following inclusion criteria: a) they have a minimum width and height of 600 pixels by 400 pixels to align picture quality with the quality of the images contained in the Medetec dataset; and b) they complement the images in the Medetec dataset where specific categories do not exist or are poorly represented. The resulting 858 images are augmented using the Python Augmentor tool to flip, scale, tilt, skew, and rotate each image. Each image is resized with a fixed ratio of 1024 by 1024 to match the input resolution of the network. The final dataset was analysed by an NHS District Nurse with expertise in pressure ulcer categorisation, and each image was tagged as one of the six classes.  On completion of the tagging process a total of 5084 objects were tagged in 4290 images: 685 tags for category I, 1401 tags for category II, 432 tags for category III, 740 tags for category IV, 899 for DTI and 927 for unstageable. Figure 1 shows the class distribution.
\begin{figure}[htp] 
	\centering
	\includegraphics[width=0.80\linewidth]{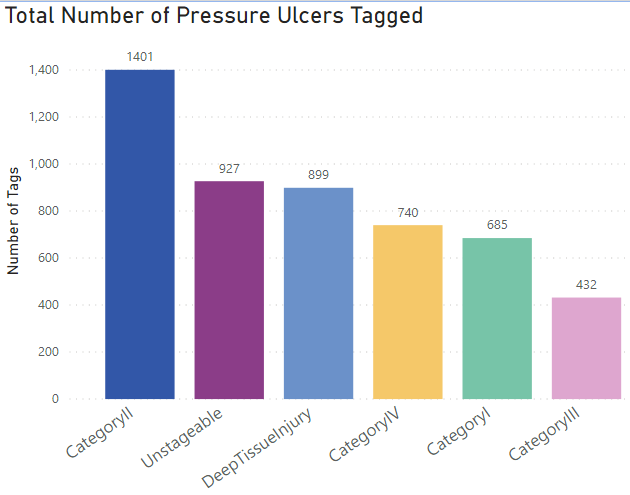}
	\caption{Class Distributions for the Tagged Dataset}
	\label{fig1} 
\end{figure}
The tagging of data was undertaken using Labelme. Binding boxes are placed around objects to identify regions of interest. All of the tagged regions in each image were exported as Extensible Mark-up Language (XML) in TensorFlow Pascal VOC format. The generated XML is converted into Comma Separated Values (CSV) using Pandas and XML. A script was used to derive both the training and validation sets based on the tagged classes. Using TensorFlow 2.2 and Pillow, both the XML and image data are converted into TFRecords for training. 
\subsection{Faster Region-Based Convolutional Neural Network}
The Faster R-CNN architecture is implemented to perform object detection and classification on images containing pressure ulcers \cite{ren2015faster}. It has three parts: a) a CNN for classification and feature map generation, b) a region proposal network (RPN) for generating Regions of Interest (RoI), and c) a regressor, which is used to find the locations of each object and its classifications. Figure 2 provides an overview of the network architecture. 
\begin{figure}[htp] 
	\centering
	\includegraphics[width=0.60\linewidth]{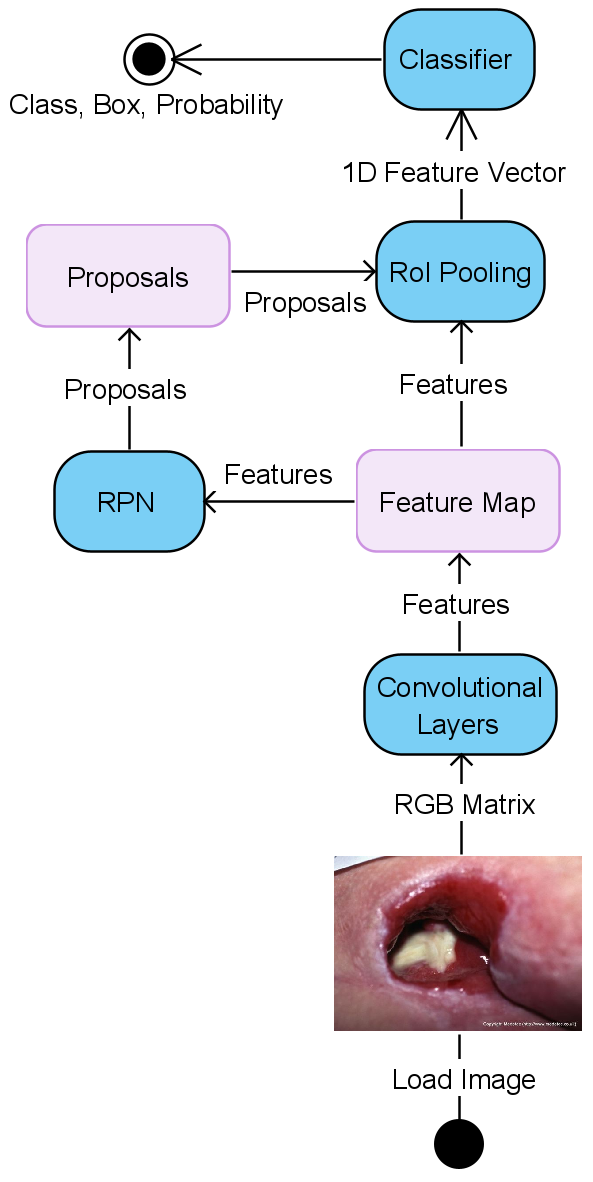}
	\caption{Faster R-CNN}
	\label{fig1} 
\end{figure}
The RPN identifies candidate pressure ulcer categories in photographs using previously learnt features in the base network (ResNet101 in this instance). The RPN replaces the selective search approach used in early R-CNN networks where region proposals were input at the pixel level rather than the feature map level. The RPN finds bounding boxes in the image with different sizes and aspect ratios as shown in figure 3. 
Anchors (fixed bounding boxes) placed throughout the image are represented using 9 different size and aspect ratio configurations. These are referenced by the RPN when first predicting object locations. The RPN is implemented as a CNN, with the feature map provided in the base network. A set of anchors for each of the points in an image are created. Note that the feature map dimensions are the same as those in the original image.
\par
The RPN generates two outputs for each anchor bounding box a) a probability objectness score and b) a set of bounding box coordinates. The first output is a binary classification, the second a bounding box regression adjustment. During the training process, all the classified anchors are placed into one of two categories a) foreground: anchors that overlap the ground-truth object with an Intersection over Union (IoU) bigger than 0.5, or b) background: anchors that do not overlap any ground truth object or have less than a 0.1 IoU with ground-truth objects. The IoU is defined as:
\begin{equation*} IoU=\frac{Anchor \ box\cap Ground \ Truth \ box}{Anchor \ box\cup Ground \ Truth \ box} \tag{1} \end{equation*} 
Anchors are randomly sampled to create mini-batches with 256 balanced foreground and background anchors. Each batch is used to calculate the classification loss using binary cross-entropy. Anchors marked as foreground in the mini-batch are used to calculate the regression loss and the correct $\Delta$ to transform the anchor into the object. If no foreground anchors are found foreground anchors are selected that have the greatest IoU with overlapping ground truth objects. This ensures that foreground samples and targets are provided for the network to learn from rather than having no anchors at all.
\par
Anchors will overlap; therefore, proposals will also overlap on the same object. Non-Maximum Suppression (NMS) is performed to delete intersecting anchor boxes with lower IoU values. IoU values greater than 0.7 describe positive object detection and values less than 0.3 describe background objects. Cation is required when setting the IoU threshold as setting it to low will result in proposals for objects being missed; too high and there will be too many proposals for the same object. It is typical to use 0.6 for the IoU threshold. The top \textit{N} proposals, sorted by score, are selected after applying NMS.
\begin{figure}[htp] 
	\centering
	\includegraphics[width=0.88\linewidth]{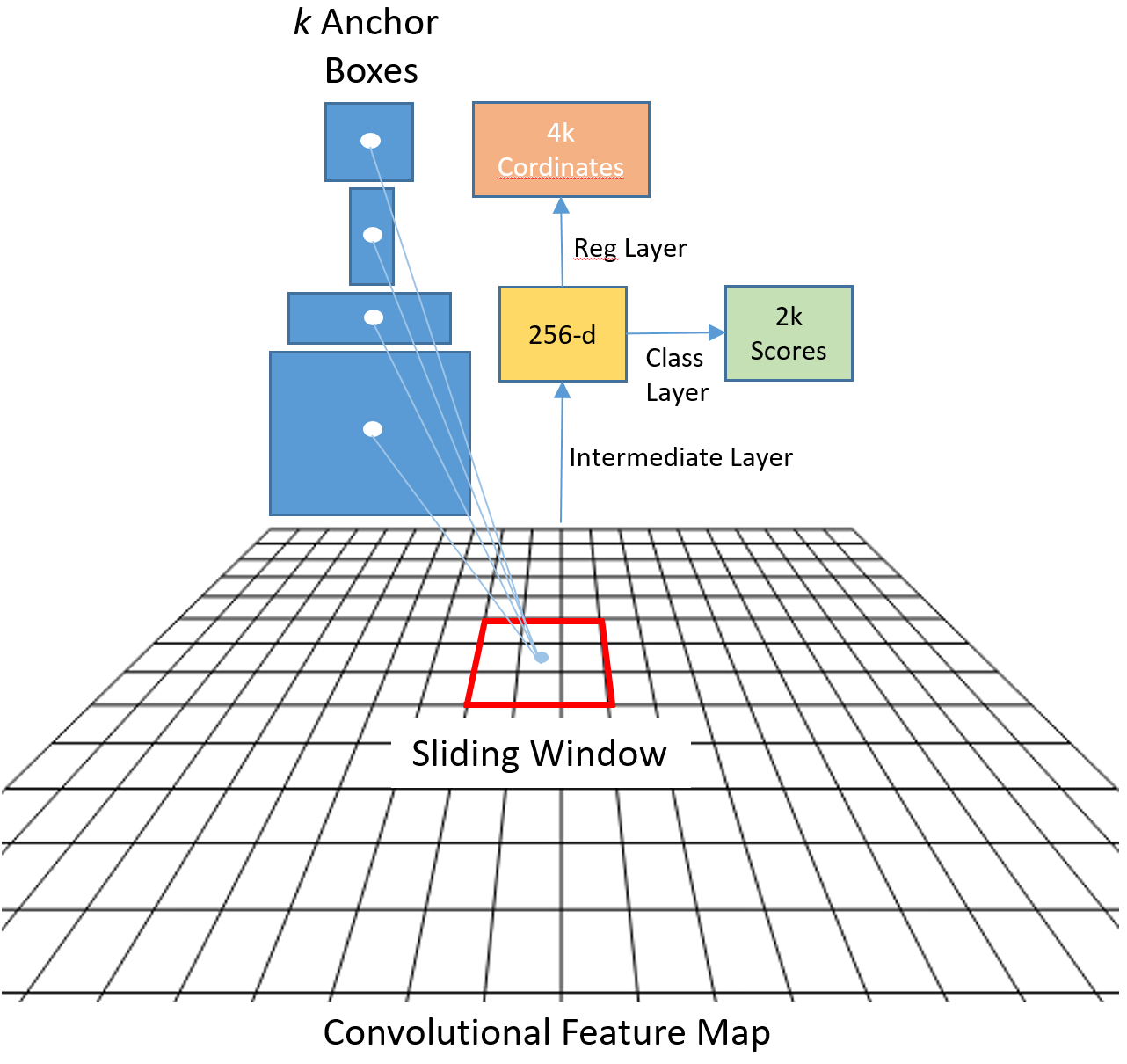}
	\caption{Region Proposal Network}
	\label{fig2} 
\end{figure}
The loss functions for both the classifier and bounding box calculation are defined as:
\begin{align*} & L_{cls}(p_{i}, p_{i}^{\ast})=- (p_{i}^{\ast}log(p_{i})+(1-p_{i}^{\ast})log(1-p_{i})) \tag{4}\\ & L_{reg}(t_{i},t_{i}^{\ast})= \Sigma_{i\in\{x,y,w,h\}}smooth_{L1}(t_{i}-t_{i}^{\ast}) \tag{5} \end{align*}
where 
\begin{equation*} smooth_{L1}(t_{i}-t_{i}^{\ast})=\begin{cases} 0.5x^{2}& if\vert t_{i}-t_{i}^{\ast}\vert < 1\\ \vert x\vert -0.5& 0ther \end{cases} \tag{6} \end{equation*}
\textit{pi} the object possibility, \textit{ti} the 4k anchor coordinate, \textit{pi*} the ground truth label, \textit{t*} the ground truth coordinate, $L_{cls}$ the classification loss (log loss), and $L_{reg}$ the regression loss (smooth L1 loss)
\par
Once the RPN step has completed there will be a set of object proposals. At this stage, the proposals do not have a class assigned to them. Each bounding box must be classified and assigned a category. In the Faster R-CNN implementation, the convolutional feature map is cropped using each proposal. Each crop is then resized to 14 * 14 * convdepth using interpolation. After cropping, max pooling with a 2x2 kernel is used to get a final 7 * 7 * 512 feature map for each proposal (via RoI Pooling). These dimensions are default parameters set by the Fast R-CNN; however, they are customizable depending on second stage use.
\par
The Fast R-CNN takes the $7*7*512$ feature map for each proposal, flattens it into a one-dimensional vector and connects it to two fully-connected layers of size 4096 with Rectifier Linear Unit (ReLU) activation. An additional fully-connected layer to identify object classes is implemented where \textit{N} describes the total number of classes and \textit{+1} the background. In parallel, a second fully-connected layer with \textit{4N} units is implemented for bounding box regression prediction. The 4 parameters correspond to $\Delta_{center_x}$, $\Delta_{center_y}$, $\Delta_{width}$, $\Delta_{height}$ for each of the \textit{N} possible classes. Figure 4 describes the Fast R-CNN architecture. 
\begin{figure}[htp] 
	\centering
	\includegraphics[width=0.88\linewidth]{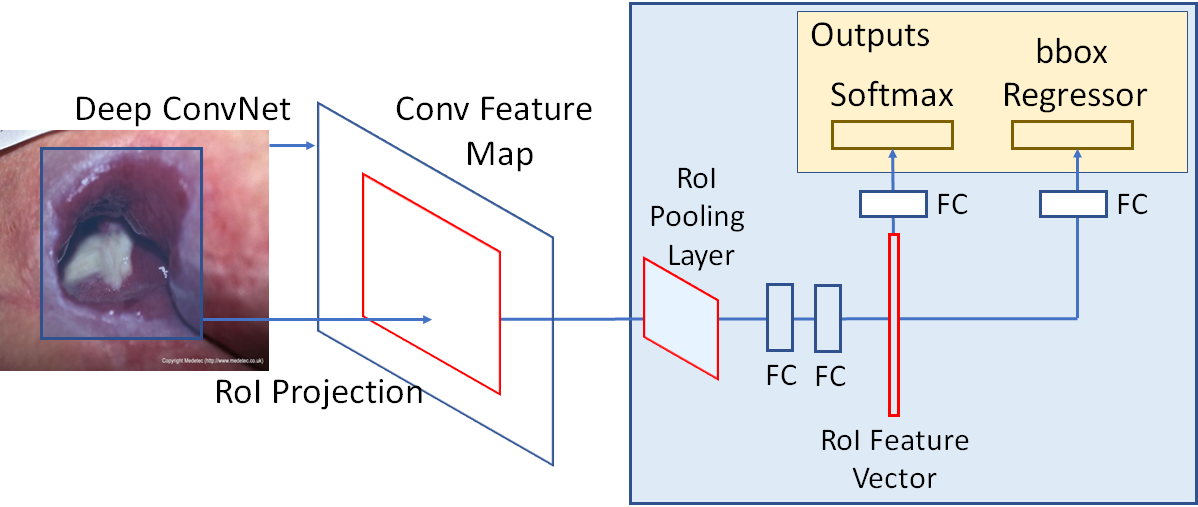}
	\caption{Fast R-CNN}
	\label{fig2} 
\end{figure}
Targets in a Fast R-CNN are calculated in a similar way to the RPN targets but with different possible classes taken into account. Proposals and ground-truth boxes are used to calculate the IoU between them. Proposals with an IoU greater than 0.5 when compared with any ground truth box get assigned to that ground truth. Proposals with an IoU between 0.1 and 0.5 are assigned to the background. Proposals with no intersection are ignored. Targets for bounding box regression can then be calculated by determining the offset between the proposal and its corresponding ground-truth box. Note this only happens for proposals that have been assigned a class based on the IoU threshold. The Fast R-CNN is trained using backpropagation and Stochastic Gradient Descent. Calculating the loss function in Fast R-CNN is defined as:
\begin{equation*} L(p,u,t^{u},\ v)=L_{cls}(p,u)+\lambda.\ [\mathrm{u}\geq 1]L_{reg}(t^{u},\nu) \tag{7} \end{equation*}
Where \textit{p} describes the object possibility, \textit{u} the classification class, \textit{t} the ground truth label, \textit{v} the ground truth coordinates for class \textit{u}, \textit{$L_{cls}$} the Loss function for classification, \textit{$L_{reg}$} the Loss function for the bounding box regressor, and $\theta$ the balancing parameter. The \textit{$L_{cls}$} is defined as:
\begin{equation*} L_{cls}(p,u)=-\log(\frac{e^{p_{u}}}{\Sigma_{j=1}^{K}e^{p_{j}}}) \tag{8} \end{equation*}
Where \textit{p} is the object possibility, \textit{u} the classification class, \textit{$L_{cls}$} the Loss function for classification and \textit{K} the number of classes. \textit{$L_{reg}$} can be calculated using the equation described in 5 with \textit{$t^{u}$} and \textit{v} as input.
\par
Following object classification, bounding box adjustments are performed. This is achieved by taking into account the class with the highest probability for that proposal. Proposals that have a background class assignment are ignored. Using the final set of objects class-based NMS is applied and, to minimise the final set of objects returned, a probability threshold is set.
\par
Putting the complete model together there are two losses for the RPN and two for the R-CNN. The four losses are combined using a weighted sum to give classification losses more weight relative to regression losses, or give R-CNN losses more power over the RPNs’.
\subsection{Transfer Learning}
Transfer learning is adopted to fine-tune a pre-trained model using the six pressure ulcer classes in our dataset. This is an important technique as training CNNs on small datasets (which we have in this study) leads to extreme overfitting due to low variance. The base model is the residual neural networks-101 (Resnet101) model \cite{he2016deep}. It has been pre-trained using the COCO dataset which contains 330 thousand images and 1.5 million object instances. Residual neural networks are deep neural networks based on a highway networks architecture \cite{srivastava2015highway}. They accelerate training in very deep neural networks and using skip connectors, avoid vanishing and exploding gradients. 
\subsection{Model Training}
Model training is performed on an HP ProLiant ML 350 Gen 9 Server with x2 Intel Xeon E5-2640 v4 series processors, 768GB of RAM and four NVidia Quadro RTX8000 graphics cards with a combined 192GB of GPU memory. TensorFlow 2.2, TensorFlow Object Detection API, CUDA 10.2 and CuDNN version 7.6 are used in the training pipeline. In the TensorFlow pipeline.config file the following hyper parameters are set:
\begin{itemize}
	\item To maintain aspect ratio resizer minimum and maximum coefficients are set to 1024x1024 pixels respectively. This minimises the scaling effect on the acquired data. 
	\item The default setting for the feature extractor coefficient is retrained to provide a standard 16-pixel stride length to maintain a high-resolution aspect ratio and improve training time. 
	\item The batch size coefficient is set to thirty-two to maintain GPU memory limits.
	\item The learning rate is set to 0.0004 to prevent large variations in response to the error.
\end{itemize}
The Adam optimizer is implemented in Resnet 101 to minimise the loss function \cite{kingma2014adam}. Unlike optimisers that maintain a single learning rate (alpha) throughout the entire training session (stochastic gradient descent), Adam calculates the moving average of the gradient $m_{t}$/squared gradients $v_{t}$ and the parameters beta1/beta2 to dynamically adjust the learning rate. Adam is defined as:
\begin{equation} 
	\begin{aligned}
		m_{t} = \beta_{1}m_{t} - 1 + (1- \beta_{1})g_{t} \\
		v_{t} = \beta_{2}v_{t} - 1 + (1 - \beta_{2})g_{t}^{2}
	\end{aligned}
\end{equation} 
where $m_{t}$ and $v_{t}$ are estimates of the first and second moment of the gradients. Both $m_{t}$ and $v_{t}$ are initialised with 0's. Biases are corrected by computing the first and second moment estimates:
\begin{equation} 
	\begin{aligned}
		\hat{m}_{t} = \frac{m_{t}}{1-\beta_{1}^{t}} \\
		\hat{v}_{t} = \frac{v_{t}}{1-\beta_{2}^{t}}
	\end{aligned}
\end{equation} 
{\parindent0pt 
	Parameters are updated using the Adam update rule:\\	
}
\begin{equation} 
	\begin{aligned}
		\theta_{t+1} =\theta_{t} - \frac{n}{\sqrt{\hat{v}_{t} + \epsilon}}\hat{m}_t\cdot
	\end{aligned}
\end{equation}
The ReLU activation function is adopted to overcome the saturation changes around the mid-point of their input which is a common problem with sigmoid or hyperbolic tangent (tanh) activations \cite{nair2010rectified}. ReLU is defined as:
\begin{equation} 
	\begin{aligned}
		g(x) = max(0,x)
	\end{aligned}
\end{equation}
\subsection{Inference Pipeline}
The trained pressure ulcer model is hosted using TensorFlow Serving. District nurses in the study use iOS and Android mobile devices over 4/5G communications to transmit photographs of pressure ulcers through a WordPress web interface hosted on Apache. Photographs received server-side are submitted to the model in TensorFlow Serving using gRPC, classified and then stored in a backend MySQL database on the server as byte code. The classification results can be viewed by clinicians in the gallery 2-3 seconds after the initial photograph is taken. Inference is undertaken on a custom-built server containing an Intel Xeon E5-1630v3 CPU, 64GB of RAM and an NVidia Tesla T4 GPU. TensorFlow 2.2, CUDA 10.2 and CuDNN 7.6 are used to inference the model. 
\subsection{Clinical Trial Protocol}
Mersey Care NHS Foundation Trust identified 50 patients for the trial (full NHS IRAS and HRA Ethical approval was obtained). Participation was based on the following inclusion criteria: a) patients were diagnosed with one or more pressure ulcers (category I, II, III, IV, DTI, and unstageable), and b) patients had the capacity to consent. One hundred and fifty District Nurses across the Merseyside region were recruited into the study. All nurses used 4/5G trust phones registered in the Trusts Mobile Device Management System (MDMS). Mobile phones connect to secure pressure ulcer management web services hosted in a data centre at Liverpool John Moores University (LJMU). Figure 5 shows the four main screens the system provides. The first screen is the landing page, the second is the user login page, the third is the main dashboard where nurses can upload and view images and finally the fourth is the upload page nurses use to take photographs of pressure ulcers and send them to LJMU servers for classification.
\begin{figure}[htp] 
	\centering
	\includegraphics[width=0.95\linewidth]{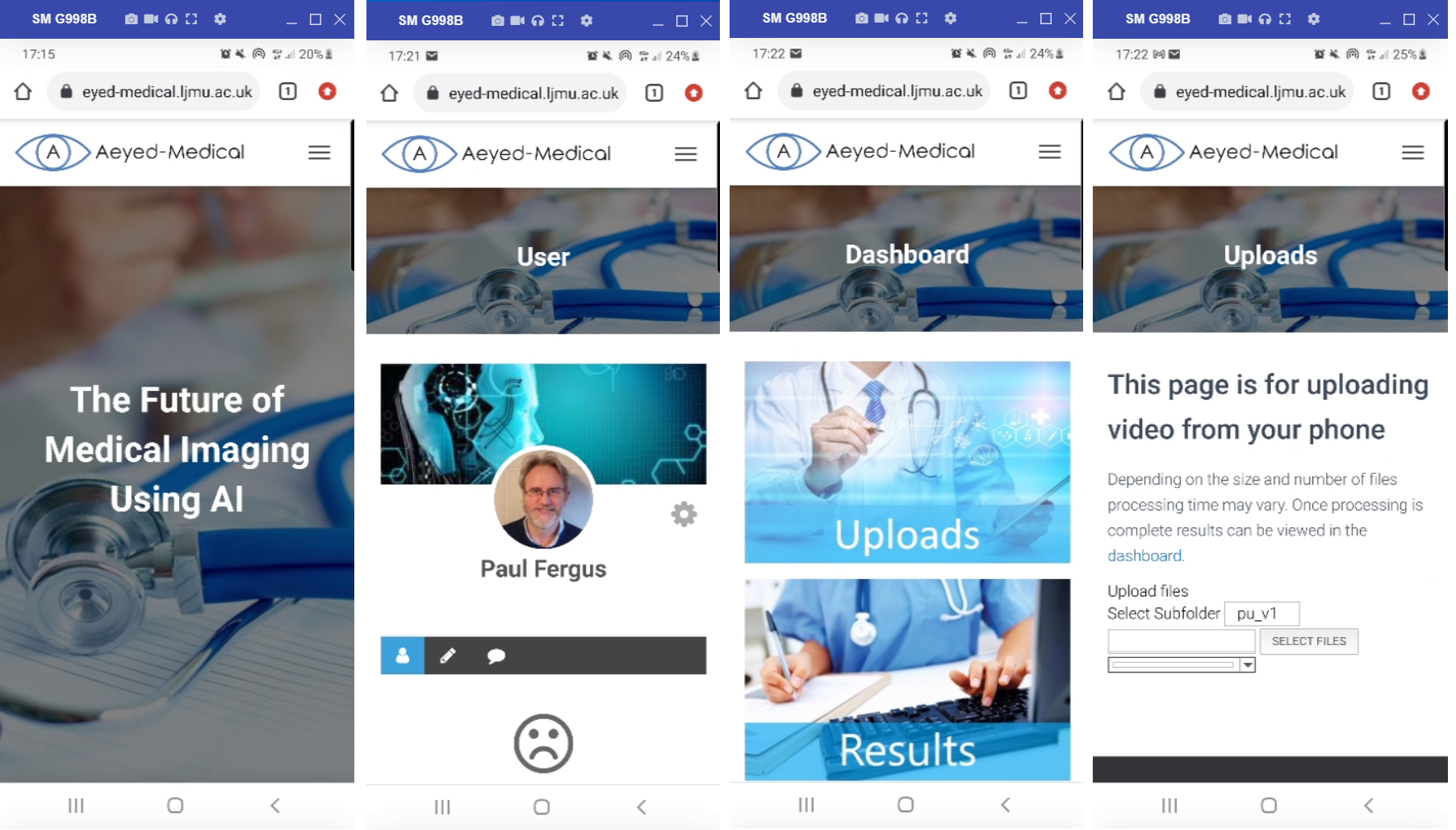}
	\caption{PUMS Mobile Phone Web Services Interface}
	\label{fig2} 
\end{figure}
Taking photographs did not impact service users or district nurses beyond normal clinical practice. The study ran between the 15th of March 2021 and the 21st of December 2021. Throughout the trial, specialist nurses reviewed the classifications made and either confirmed the category(s) was correct or reported what the correct category should be. Poor quality images, images with patient or nurse identifiable information, and images that did not contain pressure ulcers were removed from the study.    
\subsection{Evaluation Metrics}
The model’s performance during training is evaluated using RPNLoss/objectiveness, RPNLoss/localisation, BoxClassifierLoss/classification, BoxClassifierLoss/localisation and TotalLoss. These metrics are collected from Tensorboard 2.6. The RPNLoss/objectiveness measures how well the model can generate suitable bounding boxes and categorise them as either a background or foreground object. RPNLoss/localisation measures how well the RPN is at generating bounding box regressor coordinates for foreground objects. In other words, how far each anchor target is from the closest bounding box. BoxClassifierLoss/classification measures the output layer/final classifier loss and describes the computed error for prediction. BoxClassifierLoss/localisation measures the performance of the bounding box regressor. All these measures are combined to produce a total loss metric.
\par
The validation set during training is measured using mAP (mean average precision), which is a standard metric for evaluating the performance of an object detection model. mAP is defined as:
\begin{equation} 
	\begin{aligned}
		mAP = \frac{\sum_{q=1}^{Q} AveP(q)}{Q}
	\end{aligned}
\end{equation}
Where Q is the number of queries in the set and $AveP(q)$ is the average precision ($AP$) for a given query $a$.
\par
The mAP is calculated on the binding box locations for the final two checkpoints. IoU thresholds @.50 and @.75 are used to assess the overall performance of the model. This is achieved by measuring the percentage ratio of the overlap between the predicted bounding box and the ground truth bounding box and is defined as:
\begin{equation} 
	\begin{aligned}
		IoU = \frac{Area of Overlap}{Area of Union}
	\end{aligned}
\end{equation}
A threshold of @.50 measures the overall detection accuracy while the upper threshold of @.75 measures localisation accuracy.
\par
Using the final trained model, inference is measured using photographs taken during the clinical trial to evaluate the performance of the model in a real-world situation. Inference is evaluated using Precision, Recall, F1-Score and Support. Precision is defined as:
\begin{equation}
	\begin{aligned}
		Precision = \frac{TP}{TP + FP}
	\end{aligned}
\end{equation}
Recall is defined as:
\begin{equation} 
	\begin{aligned}
		Recall = \frac{TP}{TP + FN}
	\end{aligned}
\end{equation}
F1 Score is defined as:
\begin{equation} 
	\begin{aligned}
		F1 Score =2 * \frac{Precision * Recall}{Precision + Recall}
	\end{aligned}
\end{equation}
Support is used to describe the number of samples of the true response that reside within specific classes in the test set (the number of pressure ulcers in images obtained from the clinical trial). 
\par
The ground truths for images taken by nurses during the trial are provided by clinical staff at Mersey Care NHS Foundation Trust and used to calculate the detections generated by the in-trial model. Precision, Recall, F1-Score and Support are calculated with IoU@.50 for all experiments and confidence scores (CS) @.30, @.50, @.75, @.90. These metrics are used to provide an overall assessment for each class in the model during clinical trial inference. The experiments also report all false positives that reside outside of the IoU@.50 threshold at each of the four CS thresholds. The precision-recall receiver operator curve (ROC) is used to visually represent the cutoffs and the area under the curve (AUC).  
\section{Evaluation}
The results obtained during the training of the Faster RCNN model are presented first. This evaluation is then followed by two additional evaluations to determine how well the trained model performs in a clinical setting. The first evaluates the model’s ability to classify pressure ulcers in the photographs taken by district nurses. The second evaluates the same photographs cropped to only include the pressure ulcer (to remove noise and unnecessary information and to increase the size of the pressure ulcer).
\par    
\subsection{Training Results for Model Trained on Medetec and Google Images}
In the first experiment, the training set (Medetec and Google scrapped images - 4291 in total) are used to fit the model. The dataset is randomly split into training (90\%), and validation (10\%). The model is trained over 25000 epochs using a batch size of 32.
\subsubsection{Results for Training Dataset}
The results in Table 1 indicate that the model is generally good at producing candidate regions of interest (0.0593). The results also show that the RPN can effectively perform localisation on the objects identified (0.0598). The classification loss is higher (0.2015) than all other losses indicating the model is much less accurate at classifying identified objects of interest. This will correlate with the results presented for inference later in this section. In terms of box classifier localisation (0.0564), this is much more in line with the results produced by the RPN and shows that placing binding boxes around objects is not a real issue for the model. Table 1 shows the total loss (0.3770) for both the RPN and Box Classifier which is considered a good loss in object detection.
\begin{table}[htp]
	\renewcommand{\arraystretch}{1.0}
	\label{lm}
	\centering
	\caption{Tensorboard Results for Training}
	\begin{tabular}{cccc}
		\hline\hline
		Metric  							& Smoothed 	& Value		& Step 	\\ \midrule
		RPNLoss/objectness	   				& 0.0593	& 0.0521	& 25K	\\
		RPNLoss/localisation				& 0.0598	& 0.0103	& 25K	\\
		BoxClassifierLoss/classification   	& 0.2015	& 0.0622 	& 25K	\\
		BoxClassifierLoss/localisation		& 0.0564	& 0.0240	& 25K	\\
		Total Loss   						& 0.3770	& 0.1486 	& 25K	\\
		\hline\hline
	\end{tabular}
\end{table}
\subsubsection{Results for Validation Dataset}
Table 2 provides the detection boxes’ mAP metrics across several configurations. mAP provides the mean average precision over all classes averaged over IoU thresholds ranging between .5 and .95 with .05 increments. Precision (0.7743) is relatively good indicating a reduced number of false positives. The three metrics for large, medium and small objects indicate the model is better at detecting large and medium objects in images rather than smaller ones. mAP @.50IoU is the mean average at 50\% IoU and mAP @.75IoU is precision at 75\% IoU. The results in Table 2 show that the best precision values are mAP (Large)=0.8045 and mAP@.50=0.9732. Utilising large objects and the mAP@.50 IoU threshold will minimise the number of false positives returned. In other words, the validation results suggest that the model will perform reasonably well with large/medium objects and less so with smaller objects.
\begin{table}[htp]
	\renewcommand{\arraystretch}{1.0}
	\label{lm}
	\centering
	\caption{Tensor Board Results for Eval - Precision}
	\begin{tabular}{cccc}
		\hline\hline
		Metric  								& Smoothed 	& Step 	\\ \midrule
		DetectionBoxes/Precision/mAP	   		& 0.7743	& 25K	\\
		DetectionBoxes/Precision/mAP (Large)	& 0.8045	& 25K	\\
		DetectionBoxes/Precision/mAP (Medium)  	& 0.7380	& 25K	\\
		DetectionBoxes/Precision/mAP (Small)	& 0.1620	& 25K	\\
		DetectionBoxes/Precision/mAP@.50IOU   	& 0.9732	& 25K	\\
		DetectionBoxes/Precision/mAP@.75IOU		& 0.9119	& 25K	\\
		\hline\hline
	\end{tabular}
\end{table}
Table 3 provides the detection boxes mAR metrics across the same configurations used to calculate Precision. AR@1 provides the average recall with 1 detection, AR@10 is the average recall with 10 detections and AR@100 is the average recall with 100 detections. Recall in this instance represents the number of ground truths detected divided by the total number of ground truths that exist. A significant jump is seen between 1 detection and 10 detections but little change between 10 and 100. Again, the results are reasonably good when 10 or more detections are returned (0.8221 and 0.8249). In this instance, the results suggest that most of the ground truths presented were detected by the trained model. The recall values for AR@100(small, medium and large) show the average recall with 100 detections across small, medium and large objects in images. Again, the best results are obtained when large and medium objects in images are present (0.8496-0.7819) and less so for small objects (0.4212). 
\begin{table}[htp]
	\renewcommand{\arraystretch}{1.0}
	\label{lm}
	\centering
	\caption{Tensorboard Results for Eval - Recall}
	\begin{tabular}{cccc}
		\hline\hline
		Metric  								& Smoothed 	& Step 	\\ \midrule
		DetectionBoxes/Recall/AR@1	   			& 0.7308	& 25K	\\
		DetectionBoxes/Recall/AR@10				& 0.8221	& 25K	\\
		DetectionBoxes/Recall/AR@100			& 0.8249	& 25K	\\
		DetectionBoxes/Recall/AR@100 (Large)	& 0.8496	& 25K	\\
		DetectionBoxes/Recall/AR@100 (Medium)   & 0.7818	& 25K	\\
		DetectionBoxes/Recall/AR@100 (Small)	& 0.4212	& 25K	\\
		\hline\hline
	\end{tabular}
\end{table}
\subsection{Clinical Trial Results Using Trained Model}
The trained model was deployed and used in the clinical trial to analyse pressure ulcer photographs taken by district nurses during routine patient visits. During the trial, 1016 images were collected. Following quality checking, this number was reduced to 624 by removing blurry images, images that contained identifiable patient and staff information, and images that did not contain pressure ulcers. A second review was performed to remove images that were similar (the same pressure ulcer was taken repeatedly during the trial with little variance). The final test set contained 216 images (5 Category I images, 93 Category II, 11 Category III, 0 Category IV (none were seen during the trial), 30 DTI, and 77 unstageable) as shown in Figure 6. 
\begin{figure}[htp] 
	\centering
	\includegraphics[width=0.88\linewidth]{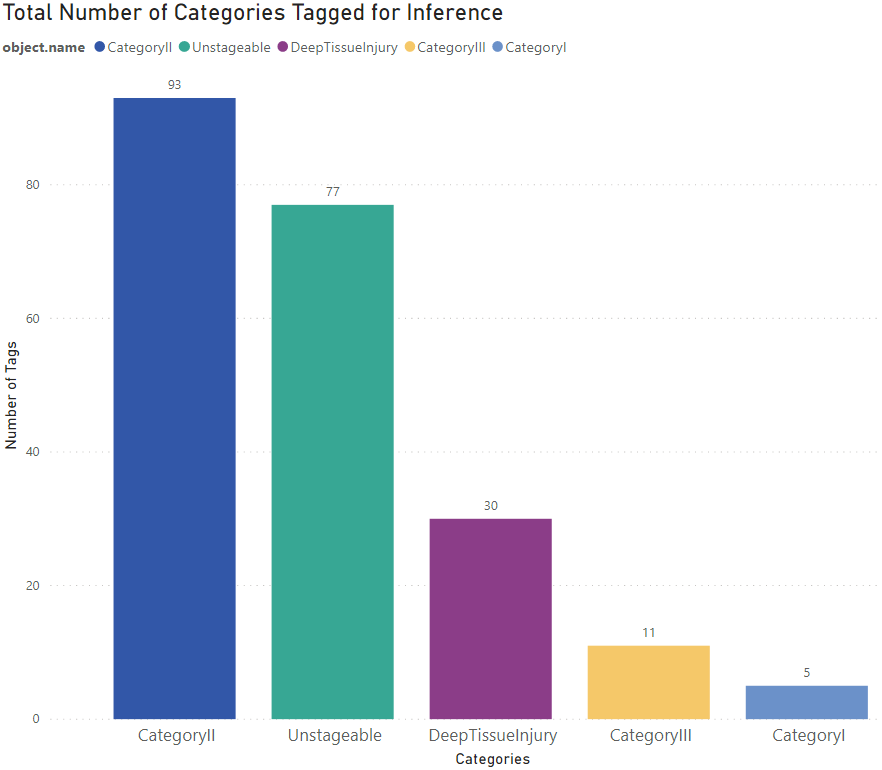}
	\caption{Inference Class Distribution}
	\label{fig5} 
\end{figure}
\subsubsection{Inference Using Uncropped Images}
In the first evaluation, the 216 photographs are evaluated class by class using Precision and Recall, with IoU@.50 and CS @.30, @.50, @.75 and @.90. The F1-score is used to calculate the harmonic mean between the Precision and Recall values. The Support for each class is also reported as is the number of false positives that reside outside of IoU@.50. Table 4 provides the performance metrics using a CS @0.30. 
\begin{table}[htp]
	\renewcommand{\arraystretch}{1.0}
	\label{lm}
	\centering
	\caption{Faster R-CNN Inference Results Using Uncropped Images with IOU@.50 CS@.30}
	\begin{tabular}{ccccc}
		\hline\hline
		Class  					& Precision	& Recall & F1-Score  & Support\\ \midrule
		CategoryI   			& 0.2222		& 0.4444 	& 0.2962	& 5  \\
		CategoryII   			& 0.3555		& 0.3609 	& 0.3581	& 93 \\
		CategoryIII   			& 0.2400		& 0.3750 	& 0.2926	& 11 \\
		CategoryIV   			& 0.0000		& 0.0000 	& 0.0000	& 0  \\
		Unstageable   			& 0.6785		& 0.4222 	& 0.5205	& 77 \\
		DTI						& 0.7619		& 0.3478 	& 0.4775	& 30 \\ \midrule
		\textbf{Mean Average}	& \textbf{0.4516} & \textbf{0.3900} & \textbf{0.3889} &	\textbf{41.8}	\\
		\hline\hline
	\end{tabular}
\end{table}
In this evaluation, 109 false positives were reported. The results overall are poor across all classes. The best performing class was unstageable - the worse was category III which is reasonable considering support was only 11. The mAP for all classes was 0.4516 and 0.3478 mAR. The F1-Score was reported as 0.3889. Increasing the CS threshold to @.50 does improve the results slightly however most Precision-Recall values are below 0.50 as shown in Table 5.  
\begin{table}[htp]
	\renewcommand{\arraystretch}{1.0}
	\label{lm}
	\centering
	\caption{Faster R-CNN Inference Results Using Uncropped Images with IOU@.50 CS@.50}
	\begin{tabular}{ccccc}
		\hline\hline
		Class  					& Precision	& Recall & F1-Score  & Support\\ \midrule
		CategoryI   			& 0.3333		& 0.4444 	& 0.3809	& 5		\\
		CategoryII   			& 0.4105		& 0.2932 	& 0.3420	& 93	\\
		CategoryIII   			& 0.3571		& 0.3125 	& 0.3333	& 11	\\
		CategoryIV   			& 0.0000		& 0.0000 	& 0.0000	& 0		\\
		Unstageable   			& 0.7200		& 0.4000 	& 0.5142	& 77	\\
		DTI						& 0.8500		& 0.3695 	& 0.5150	& 30	\\ \midrule
		\textbf{Mean Average}	& \textbf{0.5341} & \textbf{0.3639} & \textbf{0.4170} &	\textbf{41.8}	\\
		\hline\hline
	\end{tabular}
\end{table}
The @.50 threshold does however significantly reduce the number of false positives from 109 to 46. Increasing the CS further to @.75 fails to balance the Precision-Recall values and suggests that a CS @.50 is the most optimal configuration for this model as shown in Table 6.  
\begin{table}[htp]
	\renewcommand{\arraystretch}{1.0}
	\label{lm}
	\centering
	\caption{Faster R-CNN Inference Results Using Uncropped Images with IOU@.50 CS@.75}
	\begin{tabular}{ccccc}
		\hline\hline
		Class  					& Precision	& Recall & F1 Score 	& Support\\ \midrule
		CategoryI   			& 0.4285		& 0.3333 	& 0.3749	& 5		\\
		CategoryII   			& 0.4626		& 0.2330 	& 0.3099	& 93	\\
		CategoryIII   			& 0.3846		& 0.3125 	& 0.3448	& 11	\\
		CategoryIV   			& 0.0000		& 0.0000 	& 0.0000	& 0		\\
		Unstageable   			& 0.7500		& 0.3666 	& 0.4924	& 77	\\
		DTI						& 0.8750		& 0.3043 	& 0.4515	& 30	\\ \midrule
		\textbf{Mean Average}	& \textbf{0.5801} & \textbf{0.3099} & \textbf{0.3947} &	\textbf{41.8}	\\
		\hline\hline
	\end{tabular}
\end{table}
Setting the CS to @.75 reduces the false positives to 16 in line with the higher precision values but decreases the model’s ability to recall a sufficient number of ground truths. Setting the CS to @.90 further decreases the number of false positives to 4 but the overall Recall in many classes is again reduced as shown in Table 7. 
\begin{table}[htp]
	\renewcommand{\arraystretch}{1.0}
	\label{lm}
	\centering
	\caption{Faster R-CNN Inference Results Using Uncropped Images with IOU@.50 CS@.90}
	\begin{tabular}{ccccc}
		\hline\hline
		Class  					& Precision	& Recall & F1 Score	& Support\\ \midrule
		CategoryI   			& 0.7500		& 0.3333 	& 0.4615	& 5		\\
		CategoryII   			& 0.4545		& 0.1503	& 0.2258	& 93	\\
		CategoryIII   			& 0.5555		& 0.3125 	& 0.3999	& 11	\\
		CategoryIV   			& 0.0000		& 0.0000 	& 0.0000	& 0		\\
		Unstageable   			& 0.8571		& 0.3333 	& 0.4799	& 77	\\
		DTI						& 0.9285		& 0.2826 	& 0.4333	& 30	\\ \midrule
		\textbf{Mean Average}	& \textbf{0.7091} & \textbf{0.2824} & \textbf{0.4000} &	\textbf{41.8}	\\
		\hline\hline
	\end{tabular}
\end{table}
\par
{\parindent0pt 
	\textit{A) Precision-Recall Curve for Original Images @IOU.50 and @CS.50}\\	
}
The precision-recall ROC curve in Fig. 7 shows the overall model performance. The AUC values for Category I, II, III, IV, DTI and Unstageable was 0.4660, 0.6296, 0.1979, 0.0000, 0.6691 and 0.4914 respectively.
\begin{figure}[htp] 
	\centering
	\includegraphics[width=0.88\linewidth]{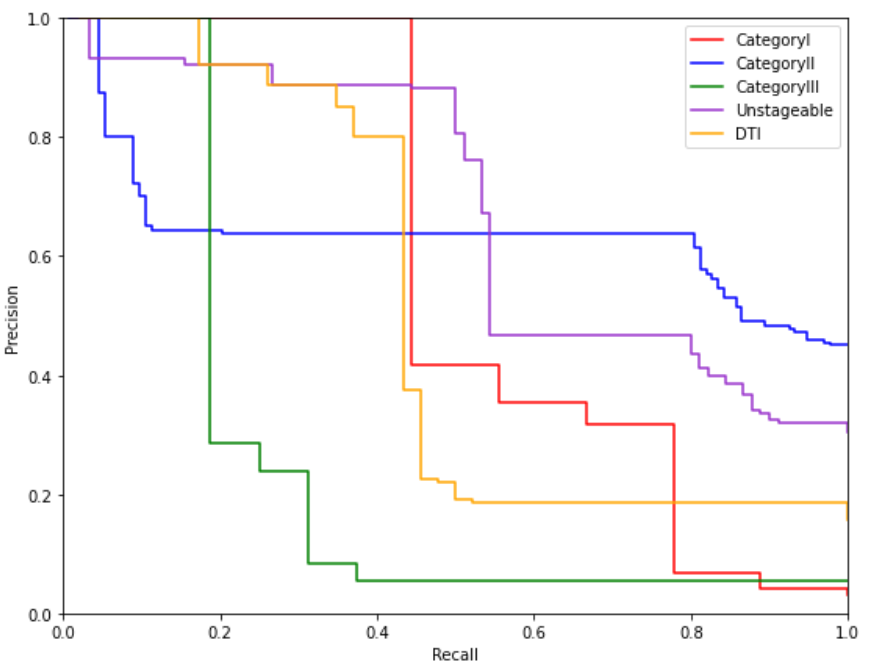}
	\caption{Medetec and Google Model with Uncropped Images.}
	\label{fig5} 
\end{figure} 
It is clear from these results the model’s performance is poor. This is partly due to the ad-hoc way in which photographs of pressure ulcers were taken. While training and best practice advice was given to district nurses, there were photograph quality issues due to poor lighting, pressure ulcer site access, and the distance mobile phones were held from the wound site when photographs were taken.
\subsubsection{Inference Using Cropped Images}
To mitigate the issues raised in the previous evaluation the 216 images were standardised by cropping them with a 1024 by 1024 aspect ratio. The assumption is that removing noise will improve the overall localisation and classification results. In other words, zoom into the image and make the pressure ulcers appear bigger. In this evaluation, the performance of the model was measured using the same metrics. Table 8 provides the results for IoU@.50 and a CS @.30.
\begin{table}[htp]
	\renewcommand{\arraystretch}{1.0}
	\label{lm}
	\centering
	\caption{Faster R-CNN Inference Results Using Cropped Images with IOU@.50 and CS@.30}
	\begin{tabular}{ccccc}
		\hline\hline
		Class  					& Precision	& Recall & F1 Score 	& Support\\ \midrule
		CategoryI   			& 0.1666		& 0.8000    & 0.2757	& 5		\\
		CategoryII   			& 0.3852		& 0.7096 	& 0.4994	& 93	\\
		CategoryIII   			& 0.1923		& 0.4545    & 0.2702	& 11	\\
		CategoryIV   			& 0.0000		& 0.0000 	& 0.0000	& 0		\\
		Unstageable   			& 0.7037		& 0.7402    & 0.7214	& 70	\\
		DTI						& 0.7037		& 0.6333    & 0.6666	& 30	\\ \midrule
		\textbf{Mean Average}	& \textbf{0.4203} & \textbf{0.6675} & \textbf{0.4866} &	\textbf{41.8}	\\
		\hline\hline
	\end{tabular}
\end{table}
Adopting this strategy improved the mean average for Precision, Recall and F1-Score but nothing significant beyond the previous set of results. The mean averages remain below 0.50 for both Precision and the F1-Score however there is a marked increase in Recall which means more of the ground truths were detected. There was an increase in the number of false positives (152). 
\begin{table}[htp]
	\renewcommand{\arraystretch}{1.0}
	\label{lm}
	\centering
	\caption{Faster R-CNN Inference Results Using Cropped Images with IOU@.50 and CS@.50}
	\begin{tabular}{ccccc}
		\hline\hline
		Class  					& Precision	& Recall & F1 Score	& Support\\ \midrule
		CategoryI   			& 0.3076		& 0.8000    & 0.4443	& 5		\\
		CategoryII   			& 0.5000		& 0.6666 	& 0.5714	& 93	\\
		CategoryIII   			& 0.2608		& 0.5454    & 0.3528	& 11	\\
		CategoryIV   			& 0.0000		& 0.0000 	& 0.0000	& 0		\\
		Unstageable   			& 0.7500		& 0.7402    & 0.7450	& 70	\\
		DTI						& 0.7307		& 0.6333    & 0.6785	& 30	\\ \midrule
		\textbf{Mean Average}	& \textbf{0.5098} & \textbf{0.6771} & \textbf{0.5584} &	\textbf{41.8}	\\
		\hline\hline
	\end{tabular}
\end{table}
Setting the CS to @.50 increased all mean average values above 0.50 as shown in Table 9. This time 93 false positives were reported. 
\begin{table}[htp]
	\renewcommand{\arraystretch}{1.0}
	\label{lm}
	\centering
	\caption{Faster R-CNN Inference Results Using Cropped Images with IOU@.50 and CS@.75}
	\begin{tabular}{ccccc}
		\hline\hline
		Class  					& Precision	& Recall & F1 Score	& Support\\ \midrule
		CategoryI   			& 0.3750		& 0.6000	& 0.4615	& 5		\\
		CategoryII   			& 0.6595		& 0.6666 	& 0.6630	& 93	\\
		CategoryIII   			& 0.5714		& 0.7272 	& 0.6399	& 11	\\
		CategoryIV   			& 0.0000		& 0.0000 	& 0.0000	& 0		\\
		Unstageable   			& 0.8378		& 0.8051 	& 0.8211	& 70	\\
		DTI   					& 0.9545		& 0.7000 	& 0.8076	& 30	\\ \midrule
		\textbf{Mean Average}	& \textbf{0.6796} & \textbf{0.6997} & \textbf{0.6786} &	\textbf{41.8}	\\
		\hline\hline
	\end{tabular}
\end{table}
With a CS @.75 the mean average results for Precision, Recall and F1-Score increase further to just below .70 as indicated in Table 10 with 45 false positives reported. 
\begin{table}[htp]
	\renewcommand{\arraystretch}{1.0}
	\label{lm}
	\centering
	\caption{Faster R-CNN Inference Results Using Cropped Images with IOU@.50 and CS@.90}
	\begin{tabular}{ccccc}
		\hline\hline
		Class  					& Precision	& Recall & F1 Score	& Support\\ \midrule
		CategoryI   			& 0.5000		& 0.6000    & 0.5454	& 5		\\
		CategoryII   			& 0.7763		& 0.6344    & 0.6982	& 93	\\
		CategoryIII   			& 0.6666		& 0.5454    & 0.5999	& 11	\\
		CategoryIV   			& 0.0000		& 0.0000    & 0.0000	& 0		\\
		Unstageable   			& 0.9384		& 0.7922    & 0.8591	& 70	\\
		DTI					   	& 1.0000		& 0.6333    & 0.7754	& 30	\\ \midrule
		\textbf{Mean Average}	& \textbf{0.7762} & \textbf{0.6410} & \textbf{0.6956} &	\textbf{41.8}	\\
		\hline\hline
	\end{tabular}
\end{table}
In the final experiment, there were further increases for precision (0.7762) and F1-Score (0.6956) however recall dropped from 0.6997 to 0.6410. As would be expected with a higher precision the number of false positives reported fell to 19. 
\par
{\parindent0pt 
	\textit{A) Precision-Recall Curve for Cropped Images @IOU.50 and @CS.50}\\	
}
The Precision-Recall ROC curve in Fig. 8 shows the model’s performance. This time the AUC values for category I, II, III, IV, DTI, and unstageable are 0.6253, 0.8552, 0.5051, 0.0000, 0.9299 and 0.8194 respectively. Compared with the results in Table 5 there was a 0.1593 improvement for category I, a 0.2256 improvement for category II, a 0.3072 improvement for category III, 0.2608 improvement for unstageable and a 0.3281 improvement for DTI. 
\begin{figure}[htp] 
	\centering
	\includegraphics[width=0.88\linewidth]{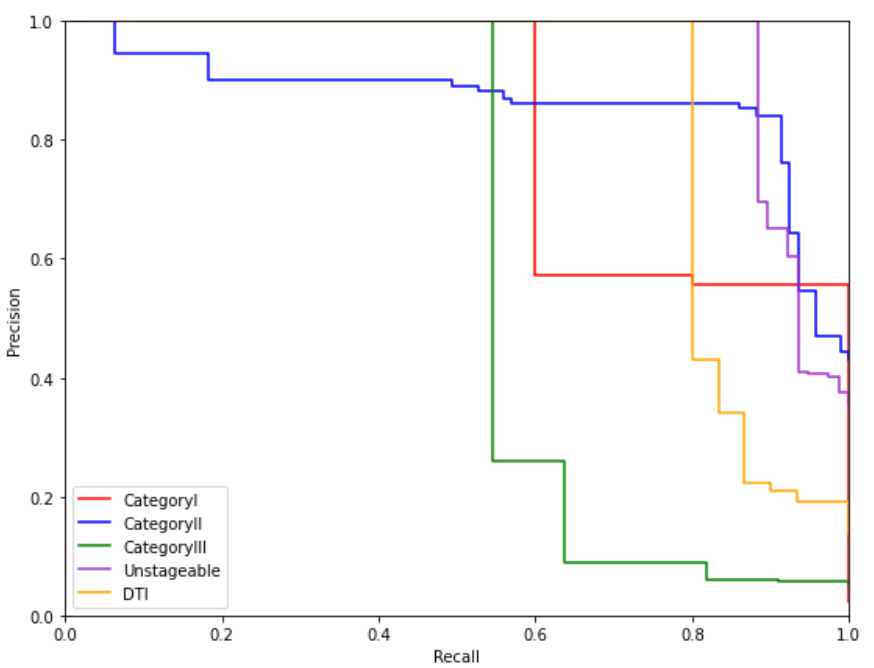}
	\caption{Medetec and Google Model with Cropped Images.}
	\label{fig6} 
\end{figure}  
\section{Discussion}
District nurses visually assess and categories pressure ulcers. This has raised concerns among healthcare professionals with regards to variability and reporting. Some practitioners will be highly adept at making assessments, others less so. The issue was addressed in this paper using DL. The results demonstrated that the Faster R-CNN using its RPN was able to effectively detect objects and apply localisation with losses of 0.0593 and 0.0598 respectively. The BoxClassifierLoss was able to produce a similar loss for localisation (0.0564) but classification loss was higher (0.2015). Collecting the required pressure ulcer images for each class (typically 1500 objects per class is required when using transfer learning) proved to be difficult in this study as there are no publicly available datasets. Images from the internet were sourced but the quality and distribution between classes was poor. For example, it was possible to collect 1401 images for category II pressure ulcers but it proved difficult to obtain enough images for all remaining categories - the worst being 432 for category III. Even with transfer learning, numbers as low as 432 have a negative impact on training and classification results. This issue negatively impacted BoxClassifierLoss for classification.
\par
The evaluation dataset showed an overall mAP of 0.7743 which is considered a good result in object detection. Table 2 also showed the mAP results for large, medium and small objects. Larger objects produced higher mAP values than smaller objects (0.8045 and 0.1620 respectively). Therefore, close up images of pressure ulcers produced better results than those taken at longer distances - a point we will return to later in the discussion. Similar results were reported for Recall as indicated in Table 3. Recall increased in line with the number of detections (AR@1 0.7308 and AR@100 0.8249) showing a strong correlation between large and small objects (0.8496 and 0.4212 respectively). Again, this suggested the model is better at recalling larger objects than smaller ones.  
\par
The trained model was evaluated in a clinical setting. The results from the first evaluation were disappointing and there are several reasons for this. First, the number of images collected from the trial was small with a significant imbalance across all classes. Category III performed the worst with an F1-Score of 0.2926. This is reasonable given that only 11 category III instances were recorded during the trial and 432 for training. The best performing category was Unstageable with an F1-Score of 0.5205. This was one of the most represented classes in the test and training datasets (77 and 927 respectively). The evaluation @.50 showed a slight increase in F1-scores across all categories except unstageable which fell slightly. The best performing category @.50 was DTI (0.5150) and the worst was category III (0.3333). Interestingly, DTI was under represented in the test set (30 objects) but had good representation in the training set (899). Category III results were as expected given that it was the least represented. Increasing the CS to @.75 and @.90 showed very little improvement in most cases, in fact, many of the categories performed worse.
\par
DTIs and unstageable (which are often larger in appearance) perform better than smaller pressure ulcers (category I and II) which are generally more difficult to analyse because of their size. This was in line with the training results discussed earlier. However, this did not fully explain the poor results. The images collected from Google and Medetec for training were pre-processed to maximise the appearance of a pressure ulcer in an image. In the trial, however, there was significant variance in how photographs were taken and the distance mobile phones were held from the wound before a photograph was taken. Larger representations of pressure ulcers were better detected (although in several instances they were miss-classified). Photographs of pressure ulcers taken at larger distances were often missed and recorded as a false negative. 
\par
To address this issue the 216 images were cropped to remove unwanted information. With the CS @.30, a marked improvement in both unstageable and DTI classifications was observed with F1-Scores of 0.7214 and 0.6666 respectively. With CS @.50 the results further with similar improvements @.75 and @.90. The best-balanced results reported was @.75 with category I equal to 0.4615, category II 0.6630, category III 0.6399, category IV 0, unstageable 0.8211 and DTI 0.8076. In comparison with the best results obtained from uncropped images (@.50 - mAP=0.5341, mAR=0.3639, mAF1=0.4170) and the results obtained @.75 in this evaluation (mAP=0.6796, mAR=0.6997, mAF1=0.6786), cropping the images significantly improved overall performance. 
\par
The results are not clinically sufficient for healthcare practice or standardised categorisation and reporting. However, they are encouraging. The model was trained on poor-quality images obtained from the Internet and despite the limitations reported, we were able to develop a pressure ulcer categorisation model and produce reasonably good results. It is hoped that this evidence will convince clinical organisations that a better model could be developed if high-quality images of pressure ulcers are openly shared with the research community. In addition, further development work should be undertaken to help clinicians photograph pressure ulcers in a standard manner. This is imperative to reduce environmental variance and improve model performance. 
\section{Conclusions and Future Work}
Pressure ulcers are a significant challenge for patients and healthcare professionals. District nurses visually assess pressure ulcers and this has raised concerns among healthcare professionals with regards to variability and miss-categorisation. While training and guidelines are given to assess, treat and report their occurrence there are inconsistencies in the type of ulcers reported, data collection and classification systems used. This paper considered this issue and reported the results from a clinical trial conducted by Mersey Care NHS Foundation Trust that evaluated the efficacy of an automated pressure ulcer categorisation and reporting system. District nurses in the study took photographs of pressure ulcers using their mobile phones and transmitted them over a 4/5G network to servers at LJMU. A total of 1016 images were collected over eight months. This number was reduced to 216 following quality checks to remove blurry images, images that contain patient or staff identifiable information, images that did not contain pressure ulcers and images that looked similar.
\par
While the results from the evaluation are encouraging the main challenge was getting access to a sufficient number of high-quality images with equal distributions across all categories. Empirically, we have found that transfer learning requires a minimum of 1500 tagged objects per class to produce results in the 90s. This means that for the six classes we would need 9000 images for training a new model. To achieve this a widespread push across all NHS trusts in the UK would be required which was beyond the scope of this study. 
\par
Nonetheless, given the challenges we believe the results highlight the benefits of DL and its ability to detect and categorise pressure ulcers. This contributes to the biomedical field and provides new insights into the use of DL and mobile platforms for pressure ulcer management and warrants further investigation. While work exists in the digital analysis of pressure ulcers using DL methods, to the best of our knowledge the study in this paper is the first comprehensive NHS clinical trial of its kind that combines DL and an enterprise mobile platform to analyse and categories pressure ulcers in near-real-time in domiciliary settings. The work builds on existing research where current methods are only capable of classifying a limited range of pressure ulcer conditions (usually the most visually distinctive). 
\par
In future work, the focus will be on obtaining NIHR funding to carry out a much larger study. There will also be a focus on understanding NHS data access policies and leveraging resources to obtain a much larger corpus of pressure ulcer images across the UK. Following sufficient imagery, we will also focus on segmentation to measure pressure ulcers and their constituent tissue types. Additional development work will be undertaken to help clinicians standardise the photography of pressure ulcers in the community to improve the predictive performance of the model.    
\section*{Acknowledgement}
The authors would like to thank the Health Research Authority (HRA) for providing ethical approval to run the study (IRAS: 253949). We would also like to thank the NIHR for adopting the study in their Clinical Research Network (CRN) portfolio and providing clinical support. This study was part-funded by The Department for Digital, Culture, Media \& Sport (DCMS) under the Liverpool 5G Create project and we would like to thank them for their support. Finally, the authors would also like to thank Mersey Care NHS Foundation Trust for running the clinical trial - in particular, Pauline Parker and William Henderson. We would also like to thank all the district nurses and the 50 patients that took part in the study and the specialist nurses that evaluated the detections generated by the AI models.
\balance
\bibliography{litrev1}
\bibliographystyle{ieeetr}

\begin{IEEEbiography}[{\includegraphics[width=1in,height=1.25in,clip,keepaspectratio]{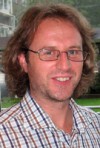}}]{Prof. Paul Fergus} is a Professor in Machine Learning at Liverpool John Moores University. Professor Fergus’s has spent just under 30 years studying Artificial Intelligence (AI) from early days of symbolic AI using Prolog and Lisp to his current research practices in Deep Learning.His research interests include machine learning for detecting and predicting preterm births as well as the detection of foetal hypoxia, electroencephalogram seizure classification and bioinformatics (polygenetic obesity, Type II diabetes and multiple sclerosis). Professor Fergus has a growing interest in conservation and is currently looking at the use of machine learning to solve different conservation-related problems. He has competitively won external grants to support his research from EPSRC, HEFCE, Royal Academy of Engineering, Innovate UK, Knowledge Transfer Partnership, North West Regional Innovation Fund and Bupa and has published over 200 peer-reviewed papers in these areas and co-authored a book entitled Applied Artificial Intelligence: Mastering the Fundamentals. Before his academic career Professor Fergus was a senior software engineer in industry for six years developing bespoke solutions for a number of large organisations.
\end{IEEEbiography}
\begin{IEEEbiography}[{\includegraphics[width=1in,height=1.25in,clip,keepaspectratio]{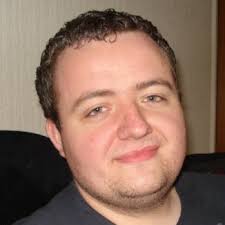}}]{Dr Carl Chalmers} is a Senior Lecturer in the Department of Computer Science at Liverpool John Moores University. Dr Chalmers’s main research interests include the advanced metering infrastructure, smart technologies, ambient assistive living, machine learning, high performance computing, cloud computing and data visualisation. His current research area focuses on remote patient monitoring and ICT-based healthcare. He is currently leading a three-year project on smart energy data and dementia in collaboration with Mersey Care NHS Trust. As part of the project a six month patient trial is underway within the NHS with future trials planned. The current trail involves monitoring and modelling the behaviour of dementia patients to facilitate safe independent living. In addition he is also working in the area of high performance computing and cloud computing to support and improve existing machine learning approaches, while facilitating application integration. 
\end{IEEEbiography}
\begin{IEEEbiography}[{\includegraphics[width=1in,height=1.25in,clip,keepaspectratio]{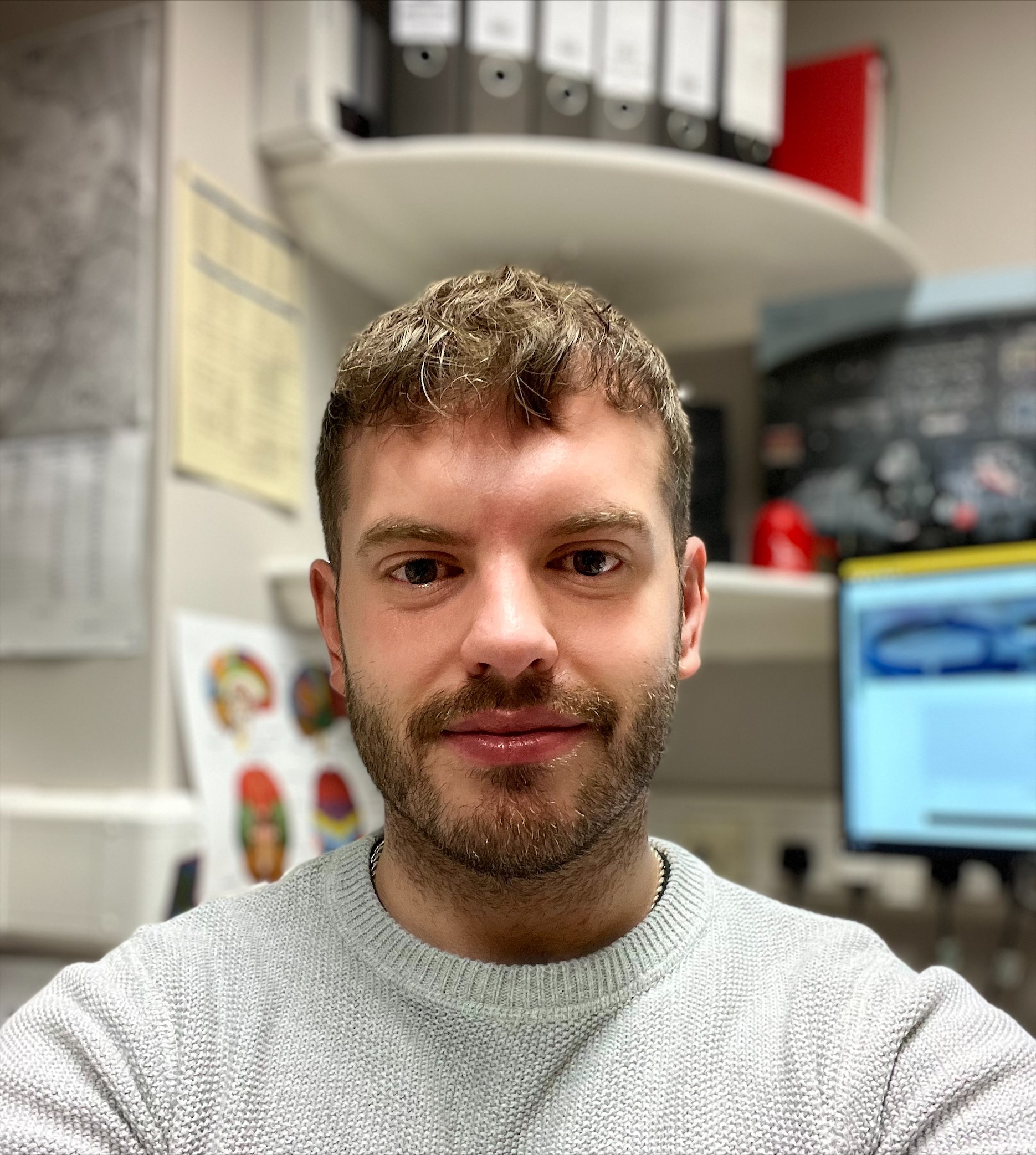}}]{Mr William Henderson} is a Clinical Research and Innovation Practitioner within Mersey Care NHS Foundation Trust. He supports both local and National Institute for Health Research (NIHR) studies. Mr Henderson is committed to delivering and supporting access to high quality research to a diverse range of patients, carers and staff. He strongly believes that research should be accessible to everyone in the NHS regardless of their mental or physical health diagnosis, helping to ensure evidence based practice is at the forefront of everything Mersey Care NHS Foundation Trust does.  
\end{IEEEbiography}

\begin{IEEEbiography}[{\includegraphics[width=1in,height=1.25in,clip,keepaspectratio]{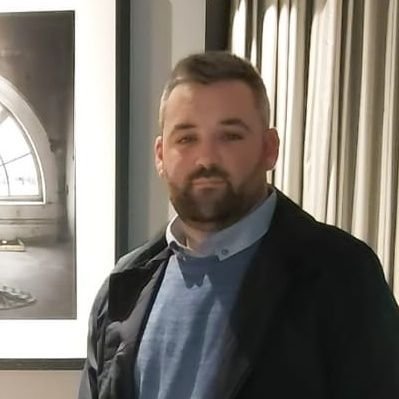}}]{Mr Danny Roberts} is the Quality Improvement Lead at the Centre for Perfect Care at Mersey Care NHS Foundation Trust. Mr Roberts has a keen research interest for improving patient safety and quality with primary car. He was a runner up in the Tissue Viability Nurse of the year category in the British Journal of Nursing awards in 2021.    
\end{IEEEbiography}

\begin{IEEEbiography}[{\includegraphics[width=1in,height=1.25in,clip,keepaspectratio]{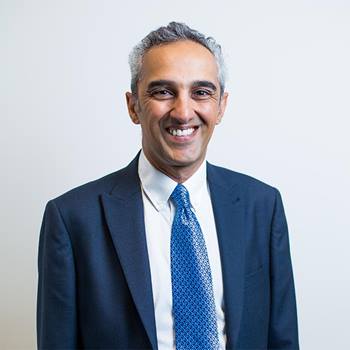}}]{Dr Atif Waraich} is Director of the School of Computer Science and Mathematics at Liverpool John Moores University (LJMU). Prior to joining LJMU (in June 2017) he was Head of Division of Digital Media and Entertainment technology at Manchester Metropolitan University (MMU) and also Enterprise Lead and was founder and Director of the Manchester Usability Laboratory. His research interests cover several areas. The first is the use of technology to enhance learning, specifically he is interested in how game like environments can be used to promote learning and to motivate learners to engage in their studies. He has a growing interest in the security and application of 5G wireless technologies. Dr. Waraich also has a research interest in the application of usability and behaviour modification techniques (including gamification) to Augmented and Virtual Reality environments and is the founder and Director of the Liverpool Immersive Experience (LIVE) Lab. He has been Principal Investigator on an Innovate UK funded project, "Rail Incident Manager. From 2011 - 2013, Dr Waraich was PI on LIFE+ Environmental Policy and Governance as part of a larger EU Project with Greater Manchester Waste Disposal Authority (GMWDA). He was lead academic at LJMU on the £5M Liverpool 5G Testbed project funded by Innovate UK and is currently lead academic on the £4M Liverpool 5G Create Programme.  He is currently Co-Investigator on the UK-Malaysia University Consortium (UK-MUC) and Digital Catalyst for Graduate Employability projects with international partners in Malaysia and Indonesia. Dr. Waraich has acted as a reviewer for numerous conferences and journals as well as a reviewer for the HEA in funding bids for technology enhanced learning. He is a member of the British Computer Society. 
\end{IEEEbiography}

\end{document}